\documentclass{article}
\usepackage[square,sort,comma,numbers]{natbib}

\usepackage[preprint]{nips_2018}
\usepackage[utf8]{inputenc} 
\usepackage[T1]{fontenc}    
\usepackage{hyperref}       
\usepackage{url}            
\usepackage{booktabs}       
\usepackage{amsfonts}       
\usepackage{nicefrac}       
\usepackage{microtype}      
\usepackage{subcaption}
\usepackage{graphicx}
\usepackage{makecell}
\usepackage[title]{appendix}
\title{Customizing an Adversarial Example Generator with Class-Conditional GANs}

\author{
  Shih-hong~Tsai\\
  Department of Electronical Engineering\\
  National Taiwan University\\
  \texttt{r05921030@g.ntu.edu} \\
}

\begin{document}
\maketitle

\begin{abstract}
 Adversarial examples are intentionally crafted data with the purpose of deceiving neural networks into misclassification. 
When we talk about strategies to create such examples, we usually refer to perturbation-based methods that fabricate adversarial examples by applying invisible perturbations onto normal data.  
The resulting data reserve their visual appearance to human observers, yet can be totally unrecognizable to DNN models, which in turn leads to completely misleading predictions.  
In this paper, however, we consider crafting adversarial examples from existing data as a limitation to example diversity. 
We propose a non-perturbation-based framework that generates native adversarial examples from class-conditional generative adversarial networks.
As such, the generated data will not resemble any existing data and thus expand example diversity, raising the difficulty in adversarial defense. 
We then extend this framework to pre-trained conditional GANs, in which we turn an existing generator into an "adversarial-example generator".
We conduct experiments on our approach for MNIST and CIFAR10 datasets and have satisfactory results, showing that this approach can be a potential alternative to previous attack strategies. 
\end{abstract}

\section{Introduction}
The term "adversarial example" first came into literature in 2013 in Szegedy's work \cite{szegedy2013}, which stated that small perturbations to neural networks' inputs would change networks' predictions. In such a case, a perturbed image of dog may be misclassified as a cat by the target network, even though the perturbations are imperceptible to humans. This flaw of deep neural networks (DNNs) allows adversaries to manipulate the models' output. Moreover, adversarial examples crafted for attacking a specific network may apply to other networks. Such transferability implies an Achilles' heel inside the mechanism of neural networks, accompanying security issues.

Ever since Szegedy et al. identified these blind spots of DNNs, related research has emerged and exploded. Attack strategies on how to create adversarial examples effectively and efficiently have been developed \cite{szegedy2013,fgsm,bim,deepfool,cw,natural,advgan}, while defense techniques on enhancing model robustness and detecting adversarial examples were proposed against \cite{towards,GPdefense,squeezing,transform}. Also, a variety of explanatory theories have been constructed and explored, attempting to identify the causes of adversarial examples \cite{szegedy2013,fgsm}. Other works investigate this phenomenon from different perspectives, including real-world application case studies \cite{bim,sitawarin2018darts} and visualization of the adversarial subspace \cite{subspace}.

As to adversarial example space, however, we found an absence among attack techniques that could enrich adversarial example diversity. Current attack strategies focus mainly on the creation of adversarial perturbations and add them back to normal data. Through this process, the visual appearance of adversarial examples still depend on existing data, which might limit the example subspace we can achieve.
We consider it as a research gap, and in this paper we study generative models that can complement those so-called “perturbation-based methods” by directly creating adversarial examples instead of generating from existing data.

\section{Related works}
In this section, first the theoretical definitions of adversarial examples are described, and then several precursors of adversarial-example generation methods such as FGSM and CW are introduced. We would also cover some GAN theories that have been combined together in order to build our method.

\paragraph{Adversarial examples}
Szegedy et al. \cite{szegedy2013} first identified this perception inconsistency between humans and machine learning models. Given an image being slightly perturbed, a DNN model may treat it as a totally different image with a completely different prediction, whereas humans may not even perceive the difference. These blind spots expose neural network models to malicious attacks, and for such slightly-changed data, they have been termed as adversarial examples. Typically, attacks of adversarial examples are mainly categorize as two types: targeted and untargeted. 

Assume that there exists a normal image $x$, the first type of attack refers to those methods that create examples $x'$ such that $C(x')=t$, given $C$, the classifier of attack target, and $t$, the class of attack target. Contrary to targeted attack, the untargeted attack urges to generate adversarial examples $x'$ such that $C(x') \neq C(x)$. In both attacks, the distance (the perturbation size) between $x$ and $x'$ should be kept below a threshold to keep them look the same. Without loss of generality, we only consider the targeted attack in this paper.

To calculate the value of perturbations, however, adversaries need to have complete access to the architecture as well as the weights of the target network, which we called a white-box attack. Conversely, for those scenarios that we cannot gain access to the target network are called a block-box setting. To deal with such black-box situations, methods have been proposed to train a substitute network that imitates the target classifier itself \cite{papernot2017practical}; white-box attacks are then applied to the substitute network.

\paragraph{Perturbation-based attack strategies}
In context of the definition of adversarial examples above, attack strategies are designed to generate minimal perturbations and to apply these perturbations back to normal data to create adversarial examples. Such methods are referred to as perturbation-based methods in this paper.

As the first proposed perturbation-based method, Szegedy et al. \cite{szegedy2013} formalizes the perturbation generation process as a optimization problem, using box-constrained L-BFGS to solve the function $f(x + r) = l$ for minimal $\| \mathbf{r} \|_{2}$, given the original image $x$, the target label $l$ and the classifier mapping function $f$. From another perspective, Goodfellow et al. proposed Fast Gradient Sign Method (FGSM) \cite{fgsm}, which is basically a gradient-descent-based algorithm aiming to increase the loss between prediction and ground-truth class for the purpose of misclassification. Based on FGSM, Kurakin et al. proposed BIM \cite{bim} to perform the gradient descent iteratively, with small perturbation at each step. Carlini and Wagner et al.  \cite{cw} also formulated this task as an optimization problem, yet they provides several techniques to make the generating process more effective, including using logit loss instead of the softmax-cross-entropy loss and converting the target to $argtanh$ space to make optimization easier. Their Carlini and Wagner's methods (abbreviated as CW's method) are now the state-of-the-art.

\paragraph{Generative Adversarial Networks (GANs)}
GANs are characterized by their powerful capability in data generation. Ian Goodfellow first introduced this adversarial framework \cite{goodfellow2014generative}, and within a few years following researchers have showed state-of-the-art performance among a wide range of applications \cite{osokin2017gans, isola2017image, zhu2017unpaired, rajeswar2017adversarial, donahue2018synthesizing}. We adopt GANs here to generate realistic adversarial examples that look alike real ones.

Moreover, to gain control over the visual class of our generated adversarial examples , we used GANs in a conditioning setting \cite{ACGAN,CGAN}. Conditional GAN (cGAN)\cite{CGAN}, as the first proposed approach of conditioning GANs, takes an extra label vector $y$ as input in addition to the original random input $z$. The class label input controls the visual class of image, whereas z provides image variability. The discriminator learns to tell the fake images $G(y,z)$ from real images $x$, while the generator learns to deceive the discriminator. Auxiliary Classifier GAN (ACGAN) \cite{ACGAN}, on the other hand, not only takes an extra input, but also gives an extra class probability output. More explicitly, the discriminator of ACGAN predicts both the image source (real or fake) and the image class, which functions as an auxiliary classifier. This modification helps stabilize training and improve image generation quality. We implemented both methods in this paper.

It is worth mentioning that concurrent work AdvGAN\cite{advgan} also utilizes GAN's generative nature to produce adversarial examples. Yet, unlike our method which directly generates adversarial instances, their methods  focus mainly on the creation of adversarial perturbations, same as traditional perturbation-based method.  
The generator of AdvGAN connects to both the discriminator and the target classifier to be attacking against. This modification to the standard GAN enables the generator to fit both the discriminator and classifier during training stage. That is, the goal of the generator is to produce perturbations to be added to normal images that can deceive the discriminator and also be misclassified by the classifier. The discriminator, on the other hand, learns to distinguish between normal and adversarial examples formed by normal data and the generated perturbations.

\section{Adaptive GAN}

\subsection{Problem Setting}
Most adversarial example generation strategies are based on developing an algorithm that extrapolates appropriate perturbations from benign data and adds them back onto those data to turn into adversarial examples. Fig. 1 illustrates the high level flow of such generation process. 
As a consequence of perturbing from existing data, the generated images resemble their unperturbed version, which is not rigorously proved yet obviously limit the adversarial diversity. To complement the vacancy of data diversity, we propose to generate adversarial examples directly from conditional generative adversarial models.

We first denote the target classifier $TC$ that adversaries aim to attack $TC: \mathbb{R}^m \rightarrow \mathbb{R}^n$ , where $m$ is the dimension of image and $n$ is the dimension of image class. Given a class label $c$ and the attack target label $t$, our goal is to train a generator $G$ that generates adversarial images $x' = G(c,z)$ such that $x'$ corresponds to class $c$ and meanwhile causes misclassification that satisfies $TC(x') = t$. 

\begin{figure}[t!]
    \centering     
    \includegraphics[height=2.2in]{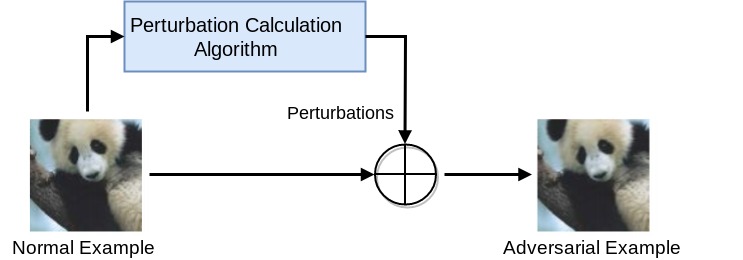}
    \caption{Illustration of a typical perturbation-based method. (The iconic panda images are adopted from the representative work of Goodfellow et al. \cite{fgsm}) }
\end{figure}

\subsection{Architecture}
The design thinking behind our model is based on an intuitive idea: imposing an additional objective on a GAN-based generator so that the generator learns to produce image samples carrying characteristics that correspond to the objective. 
We would also like the objective to be less influential, keeping the generated images as realistic as those expected from typical GANs.

To achieve our goal of adversarial attacking, we thus set the objective to minimize a certain classifier's loss w.r.t. some attack target class. Then, realistic image samples produced by the generator might be able to cause misclassification to the classifier. 

Besides, controlling the class of the generated image samples is essentially needed for targeted attack. Hence, we chose class-conditional GANs \cite{CGAN, ACGAN} as our mainframe GAN network, and impose the generator with the additional objective in expect to produce class-specified images that can mislead the target classifier.

In order to realize the objective in practice, we designed our model as a class-conditional-GAN network concatenated by the classifier of attack target, as shown in Fig.2(b). This architecture design allows us to obtain the attacking loss (which is calculated by feeding the generated images to the target classifier) and back-propagates it to generator in a straightforward manner.

At training stage, the generator takes class labels $c$, random noise $z$ and especially the label of attack target $t$ as inputs (Fig. 2(b)), contrary to conventional class-conditional GANs which only take the first two (Fig. 2(a)). Then, it is trained to minimize both the adversarial loss and attacking loss at the same time. Under this scenario, the generator acts like an adaptable animal that can adapt to external environmental constraints. Actually, it is this behavior of adaptation that we call our framework Adaptive GAN. 

The discriminator, on the other hand, is trained as conventional GANs to distinguish between fake and real images, while the target classifier stays weight-frozen. Note that our architecture is somehow similar to AdvGAN in \cite{advgan}, we think it's a coincidence on vision, and each architecture actually functions differently from one another..

After training, the generated images shall meet the following two goals: realistic for the discriminator (for humans), but misleading for the target classifier. 

Note that since we need to calculate the attack losses during training, we must have access to the target classifier architecture as well as the weights (i.e. in the white-box settings), which is the basic limitation of Adaptive GAN. 
Yet, the recent work \cite{papernot2017practical} to address the black-box attack problem by training substitute networks can be combined with our approach.

\begin{figure}[t!]
    \centering
    \begin{subfigure}[t]{0.5\textwidth}
        \centering
        \includegraphics[height=2.2in]{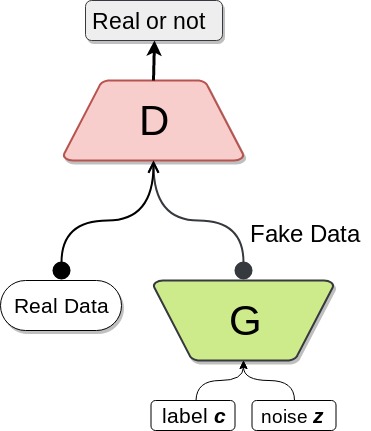}
        \caption{Class-Conditional GAN}
    \end{subfigure}%
    \begin{subfigure}[t]{0.5\textwidth}
        \centering
        \includegraphics[height=2.44in]{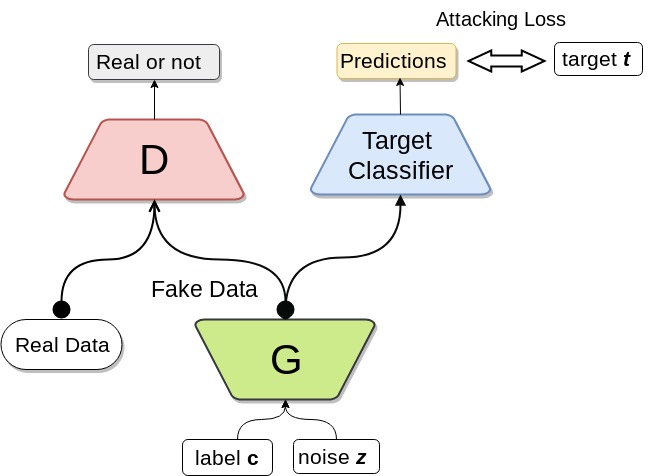}
        \caption{Adaptive GAN}
    \end{subfigure}
    \caption{The high level architecture of (a) a typical contional-GAN variant and (b) the Adaptive GAN.}
\end{figure}

\subsection{Objective}
In correspondence to the attacking loss that the target classifier contributed, an additional objective $L_{attack}$ is added accordingly:

\begin{equation} 
L_{attack} = \frac{1}{n}\sum_{n=1}^{N} I(TC(x') \neq t)  \,\cdot\, J(TC(x'), t)\label{first}
\end{equation} 
 where $I$ is an indicator function that takes the value 1 iff the predicted label $TC(x')$ does not equal to the target label $t$ and is zero otherwise, and $J$ denotes the classification loss function used for the target classifier w.r.t. the attacking target $t$ (e.g. cross-entropy).

Intuitively, $L_{attack}$ may take the value of the expected classification loss for $TC(x)$ w.r.t. to the attack target $t$:

\begin{equation} 
L_{attack}\,\,= \,\,E [ J (TC(x'), t) ]\, =\,  \frac{1}{n}\sum_{n=1}^{N}  J (TC(x'), t)  \label{two}
\end{equation} 
, which would directly reflect the attack effectiveness and is exactly Equation (\ref{first}) without the indicator function.

We observed that directly take Equation (\ref{two}) as attacking loss would make the generator fit too much on this objective. Namely, the generated images would end up in a strange shape between the given class and the target class. To alleviate this problem, we set an indicator function to mask out only the losses of images associated with unsuccessful attack, and leave successful ones less influential.
Thus, by optimizing on this adjusted objective, the generator is only trained to produce images to be misclassified as the target label, but not tuned to increase the confidence of misclassification.
Therefore, the resulting adversarial examples would end up with little distortion that is just enough to fool the target classifier. An alternative choice for balancing the loss contribution between successful attack and unsuccessful attack may be using a weighted mask with weight $\beta<1$.
We will later discuss the impact of the indicator function in the experiment in Section 4.1.

Combining $L_{attack}$ with $ L_{cGAN}$, the loss function for the mainframe conditional GAN, yields the final objective function: 
\begin{equation} 
\min\limits_{G} \max\limits_{D} L_{cGAN}(G,D) + \alpha L_{attack}(G(z,c))
\end{equation} 
 where $\alpha$ serves as the controlling parameter to balance the influence between adversarial loss and attacking loss.
Then optimization procedure for GANs can be conducted to train the generator to meet both objective: generating specified class of images, while keeping them to be misclassified as the target type.

\subsection{Turning Pre-trained Generator into Adversarial-example Generator}
Instead of training from scratch, however, the "adapting procedure" can be directly applied to pre-trained class-conditional GANs. We found that this procedure takes only a few epochs to complete, since slight changes on the output pixels is enough to cause desired misclassification.

We call it adaptive retraining that provides a way to instantly turn a pre-trained model into adversarial-example generator within few epochs of retraining, which results in slight changes to the output images given the same class condition and noise input. 
In addition, as a result of limited changes, the generative power of the original class-conditional GAN as well as the visual appearance of images produced by it are both reserved. 

Note that there are situations in which the generated adversarial examples are not as natural and convincing as real-world images, we argue that it caused by insufficient ability of GAN to describe the dataset, since the retraining only has a limited impact on the output. A more advanced model design for GAN shall be able to address this problem.

\section{Experimental results}

In this section, first we compare the effect of using different attacking loss functions in the training process. Then, we apply our method on two pubic adversarial attack challenges. 
For those experimental results hard to be quantitatively evaluated, we offer images for human assessment. Our code and models will be available upon publication.

\subsection{The Effect of Masked Loss Function}
We first discuss the effect of adding an indicator function when calculating attacking loss. The indicator function serves as a mask to filter out only the loss of unsuccessful attack (Equation \ref{first}), instead of the overall classification loss (Equation \ref{two}). We claimed that without masking, the generated adversarial examples would overfit the attack target, ending up in a shape between the original class and the target class.

To depict this scenario, we sampled "1" and "7" images targeted to be classified as "2" from models using Equation \ref{two} and Equation \ref{first} as attacking loss respectively. As can be seen in Fig. 3(a), the "1" and "7" images trained with the former would grow spots in head and tail, which makes them more like the attack target "2". On the other hand, adopting the masked loss could prevent such overfitting, as shown in Fig.3(b). 

\begin{figure}[t!]
    \centering
    \begin{subfigure}[t]{0.5\textwidth}
        \centering
        \includegraphics[height=0.8in]{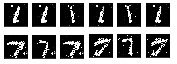}
        \caption{From models trained with naive classification loss}
    \end{subfigure}%
    \begin{subfigure}[t]{0.5\textwidth}
        \centering
        \vspace*{-2.05cm}    
        \includegraphics[height=0.75in]{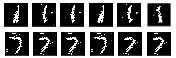}
       \vspace*{0.11cm}
       \caption{From models trained with masked loss}
    \end{subfigure}
    \caption{Comparison of the images sampled from models with attacking loss calculated using Equation 2 and Equation1 respectively. }
\end{figure}

\subsection{Adversarial Attack on MNIST}
We then applied our attack strategy on MNIST Challenge \cite{madry2017mnistchallenge}, a public challenge proposed by \cite{towards} specifically for MNIST dataset \cite{lecun1998mnist}. The challenge offers a robust classifier trained with adversarial training based on CW's method\cite{towards}, and calls for adversarial attacks that could achieve the highest attack success rate. Attackers participating in challenge are allowed to perturb each pixel of target images by at most epsilon=0.3 to prevent image distortion. 

Apparently, non-perturbation-based methods (Adaptive GAN) is hard to follow this max-norm constraint, and, in fact, should not follow such constraint by design. This makes direct comparison between two kinds of approaches a bit of unfair. However, due to lack of direct evaluation metrics, we still chose to apply our attack strategy on the challenge to see the attack effectiveness on a robustly-trained network. Also, the performance of other perturbation-based methods on leaderboard are listed for benchmarking (the performance has been converted from accuracy to successful rate).

From Table 1, we can see the highest attack success rate that Adaptive-GAN contributed to the target classifier. We consider it as a reasonable result since there is no max-perturbation constraint on non-perturbation-based method.  Yet the result still supports that our approach is effective in creating adversarial examples against robust classifiers.

To check the recognizability of adversarial examples, we randomly sampled 100 generated adversarial examples with 10 for each class of attack target in Fig. 4.
Our mainframe class-conditional model is constructed based on conditional GAN\cite{CGAN} and DCGAN\cite{dcgan}.

\begin{table}[h] 
\centering 
\caption{Attack success rate of the target classifier (MNIST Challenge)}
\label{tab:table1}
\begin{tabular}{cc} 
\toprule 
\textbf{Attack} & \textbf{Attack success rate} \\
\midrule 
\makecell[l]{100-step PGD on the cross-entropy loss \\
	with 50 random restart} & 10.38\% \\[2ex]
\makecell[l]{100-step PGD on the CW loss \\ 
	with 50 random restarts} &10.29\% \\[2ex]
\makecell[l]{100-step PGD on the cross-entropy loss	} & 7.48\% \\[1.5ex]
\makecell[l]{100-step PGD on the CW loss	} & 6.96\% \\[1.5ex]
\makecell[l]{FGSM on the cross-entropy loss} & 3.64\% \\[1.5ex]
\makecell[l]{FGSM on the CW loss} &3.60\% \\[1.5ex]
\makecell[l]{*\textbf{Adaptive GAN}} &\textbf{64.20\%} \\
\bottomrule 
\end{tabular}
\end{table}

\begin{figure}[t!]
    \centering
        \centering
        \includegraphics[height=2in]{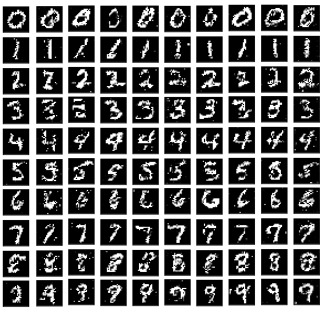}
    \caption{Adversarial examples generated from Adaptive GAN on MNIST with each row conditioned on different class label and each column conditioned on different attack target.}
\end{figure}

\subsection{Adversarial Attack on CIFAR10}

Next, we applied our attack strategy on CIFAR10 challenge \cite{madry2017cifar10challenge}, a public challenge proposed by \cite{towards} for CIFAR10 dataset \cite{cifar10}. The challenge provides a robust classifier trained with adversarial training based on CW's method, and invite adversaries to submit adversarial examples against the classifier. 

Instead of training an adversarial-example generator from scratch, we used adaptive retraining on top of a pre-trained model this time.  
By starting from a pre-trained model, we can then shorten the overall training time since CIFAR10 is a more time-consuming dataset for build GANs. 
Also, we can take the images generated from the pre-trained model as ground truth in some sense, to verify the claim the re-training would not cause too much change to the original conditional GAN's output. A pair of the images sampled from the original and retrained models is shown in Figure 5. 

Table 2 shows the attack success rate to the target classifier under different attacks, including Adaptive GAN and other perturbation-based methods on leaderboard (the performance has been converted from accuracy to successful rate). We can see that Adaptive-GAN outperforms others. Again, we have to emphasize that unlike other perturbation-based strategies, Adaptive GAN cannot limit the size of perturbation by nature, so it is not surprising that it achieves higher attack success rate than other methods. Yet the result still suggests that our method is competitive with other approaches.

As far as image quality is concerned, we have to admit that some of the generated adversarial examples are not convincing enough as those generated from perturbation-based methods. 
We think it is a practical issue for all generative models. 
Since we only provides a retraining framework which causes little change to the output of the pre-trained generative model.
And, to the best of our effort, we have constructed our model based on ACGAN\cite{ACGAN} and DCGAN\cite{dcgan} together with state-of-the-art techniques including batch normalization \cite{batch}, leaky relu as activation function, label smoothing \cite{smooth} and minibatch discrimination \cite{smooth} in this experiment.  
We think to improve the overall quality of images generated from GAN is beyond the scope of this work, and await for further research.

\begin{table}[h] 
\centering 
\caption{Accuracy of the target classifier (CIFAR10 Challenge)}
\label{tab:table1}
\begin{tabular}{cc} 
\toprule 
\textbf{Attack} & \textbf{Attack success rate} \\
\midrule 
\makecell[l]{20-step PGD on the cross-entropy loss} & 52.94\% \\[2ex]
\makecell[l]{20-step PGD on the CW loss} &52.24\% \\[2ex]
\makecell[l]{FGSM on the CW loss} & 45.08\% \\[1.5ex]
\makecell[l]{FGSM on the cross-entropy loss} & 44.45\% \\[1.5ex]
\makecell[l]{*\textbf{Adaptive GAN}} &\textbf{61.90\%} \\
\bottomrule 
\end{tabular}
\end{table}

\begin{figure}[t!]
    \centering     
    \includegraphics[height=2.2in]{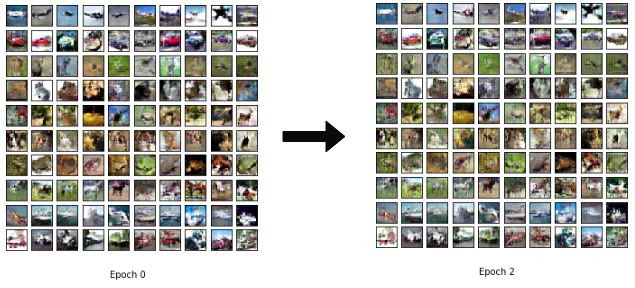}
    \caption{Adversarial examples sampled from the process of adaptive retraining on CIFAR10 Challenge, where the accuracy has been reduced by 12\% after two epochs of retraining.}
\end{figure}
\section{Conclusion}

In this paper, we study the perturbation-based nature for most adversarial-example generating methods that create examples mainly by perturbing existing data which limits data space in some manner. We then engage to complement these perturbation-based strategies 
by proposing Adaptive GAN to directly produce examples  with a modified conditional-GAN framework. Our method expands the adversarial-example subspace, since images will not have to take the appearance of existing data. We then experimented on MNIST and CIFAR10 to verify our original thinking.

A clear downside for Adaptive GAN is that it needs to be trained several times for different targets and be retrained every time when the defender modifies its model. However, we have showed in experiments that our method can apply to pre-trained GANs, turning them into adversarial-example generators. A re-training to slightly modify the pre-trained model only takes few epochs, which means the issue of training time would not limit our method.

We think the practical limitation to Adaptive GAN is that the perceptual quality of generated the adversarial examples depends heavily on the generative power of its mainframe GAN model. 
Joint effort across several GAN training techniques is required to generate acceptable quality of adversarial examples. And, in fact, the quality may not be as good when compared to those generated from traditional perturbation-based methods. In general, we still think Adaptive GAN is a promising alternative to current attack strategies since it provides a generative framework to create native adversarial examples which is irreplaceable to other methods.

\bibliographystyle{plain}
\bibliography{nips_2018}

\clearpage


\end{document}